\documentclass[conference]{IEEEtran}
\IEEEoverridecommandlockouts
\usepackage{cite}
\usepackage{amsmath,amssymb,amsfonts}
\usepackage{algorithmic}
\usepackage{graphicx}
\usepackage{textcomp}
\usepackage{xcolor}
\def\BibTeX{{\rm B\kern-.05em{\sc i\kern-.025em b}\kern-.08em
    T\kern-.1667em\lower.7ex\hbox{E}\kern-.125emX}}

\graphicspath{{figures/}}

\usepackage{pifont}
\usepackage{xcolor}
\usepackage{amsmath}
\usepackage{amsfonts}
\usepackage{amssymb}
\usepackage{caption} 
\usepackage{subcaption} 
\usepackage{bm}
\usepackage{hyperref}
\usepackage{amsthm}
\usepackage{subfiles}
\usepackage{booktabs}

\usepackage{multicol}
\usepackage{caption, subcaption}
\usepackage{graphicx}

\newcommand{\z}{\boldsymbol{z}}
\newcommand{\m}{\boldsymbol{m}}

\newcommand{\h}{\boldsymbol{h}}
\newcommand{\cmark}{\color{blue}\ding{51}}%
\newcommand{\xmark}{\color{red}\ding{55}}
\DeclareRobustCommand{\sd}[1]{\color{black!80!white}\scriptstyle #1}

\begin{document}

\title{A Variational Autoencoder for Heterogeneous Temporal and Longitudinal Data}

\author{\IEEEauthorblockN{Anonymous}
\IEEEauthorblockA{}}

\author{\IEEEauthorblockN{Mine \"{O}\u{g}retir}
\IEEEauthorblockA{\textit{Dept. of Computer Science} \\
\textit{Aalto University}\\
Espoo, Finland \\
mine.ogretir@aalto.fi}
\and
\IEEEauthorblockN{Siddharth Ramchandran}
\IEEEauthorblockA{\textit{Dept. of Computer Science} \\
\textit{Aalto University}\\
Espoo, Finland \\
siddharth.ramchandran@aalto.fi}
\and
\IEEEauthorblockN{Dimitrios Papatheodorou}
\IEEEauthorblockA{\textit{Dept. of Computer Science} \\
\textit{Aalto University}\\
Espoo, Finland \\
papat.dim@gmail.com}
\and
\IEEEauthorblockN{Harri Lähdesmäki}
\IEEEauthorblockA{\textit{Dept. of Computer Science} \\
\textit{Aalto University}\\
Espoo, Finland \\
harri.lahdesmaki@aalto.fi}
}
\maketitle

\begin{abstract}
The variational autoencoder (VAE) is a popular deep latent variable model used to analyse high-dimensional datasets by learning a low-dimensional latent representation of the data. It simultaneously learns a generative model and an inference network to perform approximate posterior inference. 
Recently proposed extensions to VAEs that can handle temporal and longitudinal data have applications in healthcare, behavioural modelling, and predictive maintenance. 
However, these extensions do not account for heterogeneous data, i.e., data comprising of continuous and discrete attributes, which is common in many real-life applications. 
In this work, we propose the heterogeneous longitudinal VAE (HL-VAE) that extends the existing temporal and longitudinal VAEs to heterogeneous data for imputing missing values and unseen time point prediction. 
HL-VAE provides efficient inference for high-dimensional datasets and includes likelihood models for continuous, count, categorical, and ordinal data while accounting for missing observations. 
We demonstrate our model's efficacy through simulated as well as clinical datasets, and show that our proposed model achieves competitive performance in missing value imputation and predictive accuracy.
\end{abstract}

\begin{IEEEkeywords}
VAE, GP, Heterogeneous Data, Time-series
\end{IEEEkeywords}

\section{Introduction}
\label{sec:introduction}

The VAE model has become very popular across scientific disciplines, especially in modern biology, where VAEs have been used to analyse single-cell RNA sequencing \cite{lopez2018deep} and microbiome data \cite{oh2020deepmicro} as well as for protein modelling \cite{hawkins2021generating}. 

While the standard VAEs are proven to be powerful generative models for complex data modelling, they assume that the data samples are independent of each other. 
To overcome this constraint, the i.i.d.\ standard Gaussian prior is replaced with a Gaussian process (GP) prior in recent work \cite{casale2018gaussian,fortuin2020gp,ramchandran2021longitudinal}. 
These GP prior VAEs can also utilise auxiliary covariates such as time, gender, and health condition. Thus, they can be used as conditional generative models. 

Most of the variations of VAEs assume that the data is homogeneous, that is, all the data is assumed to be either continuous \cite{kingma2013auto,ramchandran2021longitudinal}, binary \cite{kingma2013auto}, discrete \cite{kingma2013auto,polykovskiy2020deterministic,zhao2020variational}, or in some other form, and the data is fully observed. 
However, in various real-world problems a dataset may have several different characteristics, such as ordinal, categorical, discrete/count, and continuous data, also called heterogeneous data. Furthermore, some of the observations may be missing. 

Temporal data refers to data samples collected over time, for example, laboratory measurements of a patient over a five year period. 
On the other hand, longitudinal data is a specific type of temporal data obtained from multiple, repeatedly measured subjects, and the data contains correlations among the observations within a single subject and across multiple subjects. For example, a clinical drug trial with several patients whose lab measurements are collected on several occasions over a five year period.
Temporal and longitudinal studies are crucial in many fields, such as healthcare, economics, and social sciences, to name a few. 
In order for a VAE to be applicable to diverse real-life datasets, we require a model that: \emph{(i)} does not ignore the possible correlations between the latent embeddings across samples, \emph{(ii)} does not assume that all the features in the data are modelled using the same type of distribution, and \emph{(iii)} can handle missing observations while ensuring efficient computations. 
Some of the work on GP prior VAEs proposed approaches for modelling correlations between latent embeddings and handling missing observations \cite{fortuin2020gp,ramchandran2021longitudinal}, while Casale et al.\cite{casale2018gaussian} proposed a method to learn the auxiliary covariates from the data. 

The longitudinal variational autoencoder (L-VAE) \cite{ramchandran2021longitudinal} is an extension of the standard VAE that is tailored to longitudinal data. 
It can model both time-varying shared as well as random effects through the use of a multi-output additive GP prior. 
However, the model makes the strong assumption that the observed data is normally distributed. On the other hand, the work by \cite{nazabal2020handling} proposed a general extension to the VAE that can handle both heterogeneous and missing data; however, it is not capable of longitudinal analysis. 
In this work, we build upon these ideas and propose a model that probabilistically encodes temporal and longitudinal measurements that may comprise of discrete as well as continuous, i.e., \ heterogeneous, data onto a low-dimensional homogeneous latent space while handling missing values that are missing completely at random. 
We model the structured low-dimensional latent space dynamics using a multi-output additive GP prior as in \cite{ramchandran2021longitudinal}, and the heterogeneous data can then be reconstructed (or decoded) using a deep neural network based decoder.
\begin{figure}[t]
\begin{center}
\includegraphics[width=0.8\linewidth]{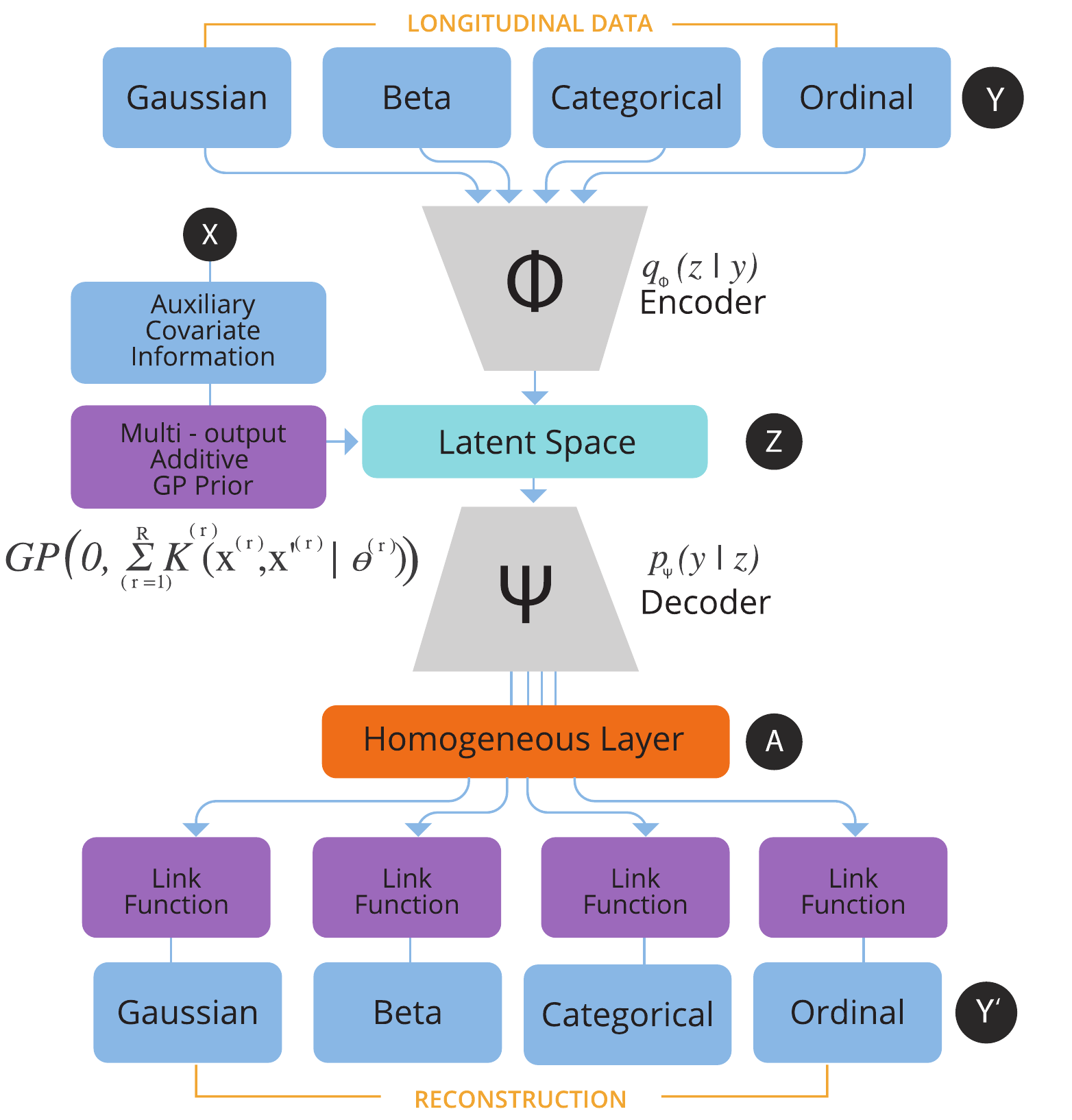}
\caption{HL-VAE overview.}
\label{fig:intro}
\end{center}
\end{figure}

Alternatively, we could use generalised linear latent and mixed models, or deep learning methods by concatenating measurements and covariates with some constraints. However, we show that our method is more expressive and intuitive.

The contributions of our work can be summarised as:
\begin{itemize}
    \item We develop a VAE, the heterogeneous temporal and longitudinal variational autoencoder (HL-VAE), for heterogeneous time-series data that models the time-dependent dynamics with auxiliary covariate information and allows for a variety of observation likelihoods. 
    \item This heterogeneous model is incorporated into an efficient inference scheme so that large and high-dimensional data can be easily modelled. 
    \item We compare our model’s performance against competing methods in missing value imputation as well as in long-term temporal and longitudinal prediction, and demonstrate the importance of assigning appropriate observation likelihoods in the heterogeneous data setting.
\end{itemize}

Our HL-VAE model is summarised in\ \autoref{fig:intro}. The source code is available at \url{https://github.com/MineOgre/HL-VAE}.
\section{Related work}
\label{sec:related-work}

\paragraph{VAEs} The variational autoencoder \cite{kingma2013auto,rezende2014stochastic} is a popular deep generative latent variable model that combines the strengths of amortised variational inference and graphical models to capture the low-dimensional latent structure of a high-dimensional, complex data distribution. However, VAEs typically assume that the latent representations are independent between the data samples and therefore fail to capture correlations between them. 
\paragraph{Gaussian process} A Gaussian process (GP) is a flexible probabilistic model that captures the covariances between the samples and is commonly used in regression as well as classification tasks \cite{rasmussen2004gaussian,williams2006gaussian}. GPs have become popular machine learning methods especially in modelling time-series data 
\cite{williams2006gaussian,roberts2013gaussian}. The Gaussian process latent variable model (GPLVM) is a GP variant that is specifically designed for embedding high-dimensional data points into a low-dimensional space \cite{lawrence2005probabilistic}. Covariate GPLVM can utilise auxiliary covariates in adjusting the embedding~\cite{martens2019decomposing}. 
\begin{table}[!t]
\caption{Comparison of related methods.} \label{table:comparison}
\centering
\resizebox{1\columnwidth}{!}{
\begin{tabular}{lccccr}
\hline \\
\textbf{Models} & \begin{tabular}[c]{@{}c@{}}\textbf{Longitudinal}\\ \textbf{study designs}\end{tabular} & \begin{tabular}[c]{@{}c@{}}\textbf{Auxiliary}\\ \textbf{covariates}\end{tabular} & \begin{tabular}[c]{@{}c@{}}\textbf{Heterogeneous}\\ \textbf{datasets}\end{tabular}  & \textbf{Minibatching} & \textbf{Reference} \\
\hline
VAE    &        \xmark     &       \xmark     &    \xmark        &     \cmark     &     \cite{kingma2013auto}\\
GPPVAE &        \xmark     &      Limited     &    \xmark     &      Pseudo    &     \cite{casale2018gaussian}\\
GP-VAE &        \xmark        &    \xmark        &   \xmark      &   \cmark    &     \cite{fortuin2020gp}\\
HI-VAE  &     \xmark      &    \xmark        &      \cmark    &       \cmark    &     \cite{nazabal2020handling}\\
L-VAE &      \cmark       &      \cmark    &       \xmark      &   \cmark    &     \cite{ramchandran2021longitudinal}\\
\hline
HL-VAE  &        \cmark     &       \cmark     &    \cmark        &      \cmark     &     Our work\\
\hline
\end{tabular}}
\end{table}
\paragraph{Models for heterogeneous data} Various probabilistic models have been developed to handle data that comprises of both continuous and discrete features.
Ramchandran et al.~\cite{ramchandran2021latent} extended the GPLVM to handle multiple likelihoods with a sampling-based variational inference technique. 
Their work enables heterogeneous datasets to be represented with low dimensional embeddings. 
Valera et al.~\cite{valera2017general} proposed a general Bayesian non-parametric latent variable model for heterogeneous datasets. 
Several authors (\cite{antelmi2019sparse,nazabal2020handling,ma2020vaem}) have introduced VAE-based methods for handling heterogeneous and incomplete data with various likelihood models.
Nazabal et al.~\cite{nazabal2020handling} proposed a model that enables the sharing of network parameters among different attributes for a single sample. Antelmi et al.~\cite{antelmi2019sparse} proposed a multi-channel approach, where each channel (each with a different likelihood) shares a common target prior on the latent representation. Ma et al.~\cite{ma2020vaem} trained individual marginal VAEs for every single variable to obtain the latent encoding for each variable and then fed them to a new multi-dimensional VAE. 
However, these models cannot capture any correlations among different samples. 
Although the models mentioned so far have not been designed for time-series or longitudinal datasets, methods that explore the domain of time-series data include, for example, the work of Moreno-Mu{\~n}oz et al. \cite{moreno2018heterogeneous}, who extend multi-output regression model for heterogeneous data. 
The approach by Gootjes et al.~\cite{gootjes2020variational} enables longitudinal data modelling for heterogeneous datasets. 
They proposed a modular Bayesian network over the latent presentations of grouped variables obtained from HI-VAE so that each group of variables can be heterogeneous. 
However, their model requires a careful design of possible graph structures. 

\paragraph{GP prior VAEs} Various works have introduced GPs to VAEs in order to overcome the i.i.d.\ assumption of the latent representations of VAEs. Casale et al.~\cite{casale2018gaussian} assumed a GP prior for the latent space of the VAE in order to incorporate the view and object information of the subjects. The GPPVAE model that they proposed assumes that the data is homogeneous (normally distributed). Moreover, its ability to model the subject-specific temporal structure is limited by the restrictive nature of the view-object GP product kernel. 
Fortuin et al.~\cite{fortuin2020gp} proposed the GP-VAE model that assumes an independent GP prior on each subject's time series. The model cannot capture temporal correlations across different subjects, and it does not utilise the auxiliary covariate information other than time.  
Ramchandran et al.~\cite{ramchandran2021longitudinal} proposed the L-VAE, which is a VAE-based model with a multi-output additive GP prior. L-VAE captures the structure of the low-dimensional embedding while utilising the auxiliary covariate information. They also proposed an efficient minibatch-based GP inference scheme. Even though L-VAE has a comprehensive capability to capture shared temporal structure as well as individual temporal structure with the help of auxiliary covariates, it assumes that the data is normally distributed, which is a major shortcoming in the modelling of longitudinal data. 
Our model addresses these shortcomings while capturing the correlations among individual subjects and temporal dependencies. We handle the various challenges of incorporating heterogeneous data while effectively building on the benefits offered by GPs and VAEs. \autoref{table:comparison} gives an overview on the relation between our approach and its most closely related methods.

\section{Background}
\label{sec:background}
 \paragraph{Notation}

In our setting, the data consists of samples, $\boldsymbol{y_i}$, and their covariate information, $\boldsymbol{x_i}$. The domain of $\boldsymbol{x}_i$ is $\mathcal{X}=\mathcal{X}_{1} \times \cdots \times \mathcal{X}_{Q}$, where $Q$ is the number of auxiliary covariates and $\mathcal{X}_{q}$ is the domain of the $q$\textsuperscript{th} covariate which can be continuous, categorical, or binary. 
The domain of $\boldsymbol{y}_i$ is $\mathcal{Y}=\mathcal{Y}_{1} \times \cdots \times \mathcal{Y}_{D}$, where $D$ is the dimensionality of the observed data. 
We assume that the variables of $\boldsymbol{y}$ are heterogeneous and that each $\mathcal{Y}_{d}$ can define a numerical domain (real, positive real, count, etc.) or a nominal domain (categorical or ordinal). 
We will define the heterogeneous domains in detail in\ \autoref{sec:hlvae}. 

A temporal dataset of size $N$ is represented with $X=\left[\boldsymbol{x}_{1}, \ldots, \boldsymbol{x}_{N}\right]^{T}$ and $Y=\left[\boldsymbol{y}_{1}, \ldots, \boldsymbol{y}_{N}\right]^{T}$, where $N$ is the number of samples taken over time for an instance (e.g.,\ an individual patient).
For a longitudinal dataset, let $P$ denote the number of unique instances (e.g.,\ individuals, patients, etc.), and $n_{p}$ denote the number of time-series samples from instance $p$. 
Therefore, the total number of longitudinal samples is $N=\sum_{p=1}^{P} n_{p}$. A pair $X_{p}=[\boldsymbol{x}_{1}^{p}, \ldots, \boldsymbol{x}_{n_{p}}^{p}]^{T}$ and $Y_{p}=[\boldsymbol{y}_{1}^{p}, \ldots, \boldsymbol{y}_{n_{p}}^{p}]^{T}$ represents the time-series data for an individual $p$, where $\boldsymbol{x}_{t}^{p} \in \mathcal{X}$ and $\boldsymbol{y}_{t}^{p} \in \mathcal{Y}$ are, respectively, the auxiliary covariate information and dependent variables. 
We denote longitudinal data as $(X,Y)$, where $X=[X_1^T,\ldots,X_P^T]^T = [\boldsymbol{x}_1,\ldots,\boldsymbol{x}_N]^T$ and $Y=[Y_1^T,\ldots,Y_P^T]^T = [\boldsymbol{y}_1,\ldots,\boldsymbol{y}_N]^T$. 
The low-dimensional latent embedding of $Y$ is given as $Z=\left[\boldsymbol{z}_{1}, \ldots, \boldsymbol{z}_{N}\right]^{T} = \left[\bar{\boldsymbol{z}}_{1},\ldots,\bar{\boldsymbol{z}}_{L} \right] \in \mathbb{R}^{N \times L}$. For example, in a clinical healthcare dataset $X$ represents the patient-specific information such as the patient identifier, age, height, weight, sex, etc. and $Y$ represents the various lab measurements.

\subsection{VAE for Heterogeneous Data}

In standard VAE models, the generative likelihood of the observed data is modelled as normally distributed. However in real-life applications, such as in healthcare, the data is typically heterogeneous. 
In order to effectively capture different data characteristics, a generative model should be able to handle various likelihood distributions. Nazabal et al.~\cite{nazabal2020handling} suggested the heterogeneous incomplete VAE model (HI-VAE) that can handle different likelihood parameters and, at the same time, capture the statistical dependencies among these data dimensions. The generative model for the $n$\textsuperscript{th} data point is
    \begin{align}
        p_{\omega}(\mathbf{y}_n, \mathbf{z}_n) &= p_{\theta}(\mathbf{z}_n) \prod_{d=1}^D p_{\psi}(y_{nd} \mid \bm{\gamma}_{nd}), \label{factorized}\\
        \bm{\gamma}_{nd} &= \bm{h}_d(\mathbf{a}_{nd}),   \quad \mathbf{A}_n = [\mathbf{a}_{n1},\ldots,\mathbf{a}_{nD}] = \mathbf{g}(\mathbf{z}_n), \nonumber
    \end{align}

where $\bm{\gamma}_{nd}$ are the likelihood parameters of the $d$\textsuperscript{th} dimension that are obtained from independent neural networks $\bm{h}_d(\cdot)$, $d \in \{1,\ldots,D\}$ and $\mathbf{A}_n \in \mathbb{R}^{S}$ is a so-called homogeneous layer where $S=s_1+\dots+s_D$ denotes the total number of dimension for each independent neural network, i.e., $s_d=|\mathbf{a}_{nd}|$. 

We can approximate the true posterior with a variational posterior as in the standard VAE by defining  $q_{\phi}\left(\mathbf{z}_{n} \mid \mathbf{y}_{n} \right)=\mathcal{N}\left(\boldsymbol{\mu}_{\phi}\left(\tilde{\mathbf{y}}_{n} \right),\boldsymbol{\Sigma}_{\phi}\left(\tilde{\mathbf{y}}_{n}\right)\right)$,  where $\boldsymbol{\mu}_{\phi}$ and $\boldsymbol{\Sigma}_{\phi}$ are the encoder networks, and $\tilde{\mathbf{y}}$ represents the encoder input with missing values replaced by zeros. In order to handle missing data in $X$, the ELBO is computed only over the observed data.
With the generative model given by\ \autoref{factorized}, the likelihoods of the data can be modelled using various distributions. 

\subsection{Gaussian Process Prior VAE}

The standard VAE model assumes that the joint distribution factorises across the samples. 
GPs, however, are well-suited to model these correlations. 
Extension of VAEs, such as GPPVAE \cite{casale2018gaussian}, GP-VAE \cite{fortuin2020gp} and L-VAE \cite{ramchandran2021longitudinal}, combine VAEs with GPs. 
These methods rely on a GP prior (instead of a standard Gaussian prior) in the latent space of a VAE to model multivariate temporal and longitudinal data. 

The standard VAE assumes a factorisable prior on the latent space $p_{\theta}(Z)=\prod_{i=1}^{N} p_{\theta}(\boldsymbol{z}_{i})$, and does not utilise the covariate information of the data. 
Meanwhile, the prior for the latent variables of the GP prior VAEs $p_{\theta}(Z \mid X)$ corresponds to a GP prior which factorises across the latent dimensions.
Specifically, for each of the $l \in \{1,\ldots,L\}$ dimensions, the latent model is $f_l(\boldsymbol{x}) \sim  \mathcal{GP}(\mu_l(\boldsymbol{x}), k_l(\boldsymbol{x},\boldsymbol{x}^{\prime}))$, where $\mu_l(\boldsymbol{x})$ and $k_l(\boldsymbol{x},\boldsymbol{x}^{\prime})$ are the mean and the covariance function (CF) of any pair of auxiliary covariates $\boldsymbol{x}$ and $\boldsymbol{x}^{\prime}$. 
We assume $\mu_l(\boldsymbol{x})=0$. Consequently, the GP prior can be written as
\begin{equation}
p_{\theta}(Z \mid X) = \prod_{l=1}^L p_{\theta}(\bar{\boldsymbol{z}}_l \mid X) =  \prod_{l=1}^L \mathcal{N}(\bar{\boldsymbol{z}}_l \mid \boldsymbol{0},\Sigma_l + \sigma_l^2 I_N),
\label{eq:gp_prior}
\end{equation}
where $\Sigma_l = K_{XX}^{(l)}$ is the $N \times N$ covariance matrix corresponding to the CF of the GP component of the $l$\textsuperscript{th} latent dimension given by $k_l(\boldsymbol{x},\boldsymbol{x}^{\prime})$.

We formulate our method using the L-VAE model which assumes additive covariance functions. L-VAE allows for more general and flexible CFs than other methods (GPPVAE and GP-VAE), and these can indeed be seen as special cases of the more general L-VAE model. L-VAE specifies the structure of the data among the observed samples with an additive multi-output GP prior over the latent space whose CF depend on the auxiliary covariates $X$, as described in \cite{ramchandran2021longitudinal}.

The generative model of L-VAE is formulated as:
\begin{align}
\begin{aligned}
p_{\omega}(Y \mid X) &=\int_{Z} p_{\psi}(Y \mid Z, X) p_{\theta}(Z \mid X) d Z \\
&=\int_{Z} \prod_{n=1}^{N} p_{\psi}\left(\boldsymbol{y}_{n} \mid \boldsymbol{z}_{n}\right) p_{\theta}(Z \mid X) d Z, \label{marginalized_LVAE}
\end{aligned}
\end{align}
where $\omega=\{\psi, \theta\}$ is the parameter set and the probabilistic decoder, $p_{\psi}\left(\boldsymbol{y}_{n} \mid \boldsymbol{z}_{n}\right)$, is assumed to be normally distributed. 
The prior for the latent variables, $p_{\theta}(Z \mid X)$, corresponds to the multi-output additive GP prior which factorises across the latent dimensions as in\ \autoref{eq:gp_prior}.
Specifically, for each of the $l \in \{1,\ldots,L\}$ dimensions the latent model is $f_l(\boldsymbol{x}) = f_l^{(1)}(\boldsymbol{x}^{(1)}) + \ldots + f_l^{(R)}(\boldsymbol{x}^{(R)})$ where $f_l^{(r)}(\boldsymbol{x}^{(r)})\sim \mathcal{GP}(\boldsymbol{0},k_l^{(r)}(\boldsymbol{x}^{(r)},\boldsymbol{x}^{(r)\prime}))$. 
$R$ refers to the total number of additive components, and each separate GP, $f_l^{(r)}(\boldsymbol{x}^{(r)})$, depends only on a small subset of covariates $\boldsymbol{x}^{(r)} \in \mathcal{X}^{(r)} \subset \mathcal{X}$. 
The covariance matrix of the GP prior in\ \autoref{eq:gp_prior} leads to $\Sigma_l = \sum_{r=1}^R K_{XX}^{(r,l)}$, where $K_{XX}^{(r,l)}$ is the $N \times N$ covariance matrix corresponding to the CF of the $r$\textsuperscript{th} GP component of the $l$\textsuperscript{th} latent dimension: $k_l^{(r)}(\boldsymbol{x}^{(r)},\boldsymbol{x}^{(r)\prime})$. 

The variational posterior of $Z$ approximating the true posterior is defined as the product of multivariate Gaussian distributions across samples,
${q_{\phi}(Z \mid Y) =\prod_{n=1}^{N} \prod_{l=1}^{L} \mathcal{N}\left(z_{n l} \mid \mu_{\phi, l}\left(\tilde{\boldsymbol{y}}_{n}\right), \sigma_{\phi, l}^{2}\left(\tilde{\boldsymbol{y}}_{n}\right)\right)}$,
where the probabilistic encoder is represented by neural network functions $\mu_{\phi, l}$ and $\sigma_{\phi, l}^2$ (parameterised by $\phi$) that determine the means as well as the variances of the approximating variational distribution. The ELBO for L-VAE as derived in \cite{ramchandran2021longitudinal}, is
\begin{align}
\label{ELBO_LVAE}
    &\log p_{\omega}(Y|X) \geq \mathcal{L}(\phi,\psi,\theta;Y,X)\\
    &\triangleq \mathbb{E}_{q_{\phi}(Z|Y)}\left[\log p_{\psi}(Y|Z) \right] - D_{\mathrm{KL}}(q_{\phi}(Z|Y)||p_\theta(Z|X)).\nonumber
\end{align}
The first term in Eq.\ \ref{ELBO_LVAE}, which is the reconstruction loss, does not involve the GP and can be obtained directly as 
\begin{align*}
\mathbb{E}_{q_{\phi}(Z \mid Y)}[\log&\ p_{\psi}(Y \mid Z)] = \\
& \sum_{n=1}^{N} \sum_{d=1}^{D} \mathbb{E}_{q_{\phi}\left(\boldsymbol{z}_{n} \mid \boldsymbol{y}_{n}\right)}\left[\log p_{\psi}\left(y_{n d} \mid \boldsymbol{z}_{n}\right)\right]. 
\end{align*}
The second term, the KL divergence between the variational posterior of $Z$ and the multi-output additive GP prior over the latent variables, $D_{\mathrm{KL}} \triangleq D_{\mathrm{KL}}(q_{\phi}(Z \mid Y)||p_\theta(Z \mid X))$, also factorises across the $L$ latent dimensions:
$D_{\mathrm{KL}} = \prod_{l=1}^L D_{\mathrm{KL}}(q_{\phi}(\bar{\z}_l \mid Y,X)||p_\theta(\bar{\z}_l \mid X))$. Each of the KL divergences are available in closed-form. However, their exact computation takes $\mathcal{O}(N^3)$. 

Ramchandran et al.~\cite{ramchandran2021longitudinal} proposed a mini-batch compatible upper bound for this KL divergence that scales linearly in the number of data points. Briefly, the method separates the individual specific random component (an additive GP component that corresponds to the interaction between instances and time) from the other additive components of the covariance function. 
Therefore (ignoring the subscript $l$ for clarity), the covariance matrix can be written as $\Sigma = K_{XX}^{(A)} + \hat{\Sigma}$, where $K_{XX}^{(A)} = \sum_{r=1}^{R-1} K_{XX}^{(r)}$ contains the first $R-1$ components, and does not have an interaction between instances and time. The last term contains the individual-specific random component $\hat{\Sigma} = \text{diag}( \hat{\Sigma}_1, \dots, \hat{\Sigma}_P )$, where $\hat{\Sigma}_p = K_{X_pX_p}^{(R)} + \sigma_z
^2 I_{n_p}$. The efficient KL upper bound is then obtained by assuming the standard low-rank inducing point approximation for $K_{XX}^{(A)}$ with $M$ inducing points $S=(\boldsymbol{s}_1,\ldots,\boldsymbol{s}_M)$, and the corresponding outputs $\boldsymbol{u} = (f(\boldsymbol{s}_1),\ldots,f(\boldsymbol{s}_M))^T =  (u_1,\ldots,u_M)^T$ (separately for each latent dimension $l$):
\begin{align}
\begin{aligned}
\label{eq:minibatchbound}
    \hat{D}_{\text{KL}} \le & \  \frac{1}{2}\frac{P}{\hat{P}} \sum_{p \in \mathcal{P}} \Upsilon_p - \frac{N}{2}  + D_{\text{KL}}(\mathcal{N}(\m,H) || \mathcal{N}(\boldsymbol{0},K_{SS}^{(A)})),
\end{aligned}
\end{align}
where the summation is over a subset of unique instances $\mathcal{P} \subset \{1,\ldots,P\}$ with $|\mathcal{P}|=\hat{P}$, $\Upsilon_p$ involves computing a quantity for instance $p$ (see \cite{ramchandran2021longitudinal}), and $\m$ and $H$ are the global variational parameters for $\boldsymbol{u} \sim \mathcal{N}(\m,H)$. For a detailed derivation, we refer the reader to \cite{ramchandran2021longitudinal}.

\section{Heterogeneous longitudinal variational autoencoder}
\label{sec:hlvae}

In this work, we extend the GP prior VAE models to handle heterogeneous data that may contain missing values. The generative model of our HL-VAE is formulated as in \autoref{marginalized_LVAE}, but makes use of a heterogeneous likelihood:
    \begin{align*}
        p_{\omega}(Y \mid X)=&\int_{Z}  \underbrace{p_{\psi}(Y \mid Z, X)}_{\text{Heterogeneous likelihood}} \underbrace{p_{\theta}(Z \mid X)}_{\text{GP prior}} d Z,
    \end{align*}
    where $\omega = \{\psi, \theta \}$ is the set of parameters, and the data is modelled with a heterogeneous likelihood:
    \begin{align}
        p_{\psi}(Y \mid Z, X) &= \prod_{n=1}^N p_{\psi}\left(\boldsymbol{y}_{n} \mid \boldsymbol{z}_{n}\right)= \prod_{n=1}^N \prod_{d=1}^D p_{\psi}(y_{nd}|\boldsymbol{\gamma}_{nd}), \nonumber \\ 
        \boldsymbol{\gamma}_{nd} &= [ l_1(\boldsymbol{h}_{1d}(\mathbf{a}_{nd})), \ldots , l_W(\boldsymbol{h}_{Wd}(\mathbf{a}_{nd}))], \label{eq:het_gamma} \\
        \mathbf{A}_n &= [\mathbf{a}_{n1},\ldots,\mathbf{a}_{nD}] = \mathbf{g}_{\psi}(\mathbf{z}_n).\label{eq:homogenous_layer}
    \end{align}
The decoder $\mathbf{g}_{\psi}$ first maps the latent code of the $n$\textsuperscript{th} sample, $\boldsymbol{z}_n$, to a homogeneous array (or layer) of outputs $\boldsymbol{A}_n \in \mathbb{R}^{U \times D}$, where $U$ is the user-chosen depth (\autoref{eq:homogenous_layer}). 
The $d$\textsuperscript{th} column of the layer, $\boldsymbol{a}_{nd}$, is then passed through $W$ independent neural networks $\boldsymbol{h}_{wd}$ (\autoref{eq:het_gamma}) where $W$ is the number of likelihood parameters for the $d$\textsuperscript{th} feature and the networks $\boldsymbol{h}_{wd}$ have parameters that are optimised. 
Finally, the outputs of the neural networks $\boldsymbol{h}_{1d},\ldots,\boldsymbol{h}_{Wd}$ are passed though link functions $l_1,\ldots,l_W$ that define the likelihood parameters $\boldsymbol{\gamma}_{nd}$ for the $d$\textsuperscript{th} feature $y_{nd} \in \mathcal{Y}_d$ with a specific likelihood model (\autoref{eq:het_gamma}). We define the link functions for a variety of likelihood models below, and our model can be easily extended to other likelihoods as well.

\textbf{Gaussian distribution:}  For the Gaussian likelihood, the likelihood parameters are given as $\boldsymbol{\gamma}_{nd} = \{l_1(\boldsymbol{h}_{1d}(\mathbf{a}_{nd})), l_2(\boldsymbol{h}_{2d}(\mathbf{a}_{nd}))\}$, where the parameters correspond to the mean and the variance respectively, and $W=2$. The link functions, $l_1(\cdot)$ and $l_2(\cdot)$, constrain the mean and the variance of the likelihood. If the data is within the interval $[0,1]$, we make use of the \textit{sigmoid} function for $l_1(\cdot)$. In all other cases, we  choose $l_1(\cdot)$ to be the identity function. The variance is constrained to be positive by choosing the \textit{softplus} function for $l_2(\cdot)$. Likelihood parameters can also be modelled as $\boldsymbol{\gamma}_{nd} = \{l_1(\boldsymbol{h}_{1d}(\mathbf{a}_{nd})), l_2(\sigma_{d})\}$ with a learnable free-variance parameter across the dataset for each variable. 

\textbf{Log-normal distribution:} For the log-normal distribution, the logarithm of the data is said to be normally distributed. 
The likelihood parameters, $\boldsymbol{\gamma}_{nd}$, are similar to the Gaussian distribution. 
We again make use of the identity function for $l_1(\cdot)$ and \textit{softplus} function for $l_2(\cdot)$.


\textbf{Poisson distribution:} The Poisson likelihood is specified by a single positive parameter (the rate parameter). The likelihood parameters are given as $\boldsymbol{\gamma}_{nd} = \{l_1(\boldsymbol{h}_{1d}(\mathbf{a}_{nd}))\}$ and $W=1$. We make use of the \textit{softplus} function for $l_1(\cdot)$ in order to constrain the parameter to non-negative values.

\textbf{Categorical distribution:} For the categorical likelihood, the data with $R$ categories is represented using one-hot encoding and we utilise a multinomial logit model. The output of the neural network represents the unnormalised probabilities for $R-1$ categories (i.e.\ except the first category). The log-probability of the first category is set to $0$. Therefore, the probability of each category $r$ is: $p_{\psi}\left(y_{n d}=r \mid \bm{\gamma}_{n d}\right)=\frac{\exp\left(-\h_{rd}\left(\mathbf{a}_{nd} \right) \right)}{\sum_{q=1}^{R} \exp\left( -\h_{qd}\left(\mathbf{a}_{nd} \right)\right)}$, where  $\boldsymbol{\gamma}_{nd} = \{0, l_1(\boldsymbol{h}_{1d}(\mathbf{a}_{nd})), \ldots, l_{R-1}(\boldsymbol{h}_{{R-1}d}(\mathbf{a}_{nd}))\}$ and all the link functions are identity functions.

\textbf{Ordinal distribution:} Assuming $R$ levels, we represent the ordinal data with the thermometer encoding and make use of the ordinal logit model. The probability for each level is computed as $p_{\psi}\left(y_{n d}=r \mid \bm{\gamma}_{n d}\right)=p\left(y_{n d} \leq r \mid \bm{\gamma}_{n d}\right)-p\left(y_{n d} \leq r-1 \mid \bm{\gamma}_{n d}\right)$ where $p\left(y_{n d} \leq r \mid \bm{\gamma}_{nd}) \right)=\frac{1}{1+\exp \left( -\left(l_r(\theta_{r}) -c_{d}\left(\mathbf{a}_{nd} \right)\right)\right)}$, and $\bm{\gamma}_{n d}=\left\{c_{d}\left(\mathbf{a}_{n d}\right)=l_1(h_{{1}d}(\mathbf{a}_{nd})), l_2(\theta_{1}), \ldots, l_2(\theta_{R-1}) \right\}$, as well as $W=1$. We choose the \textit{softplus} function as the link function. The threshold values $\theta_{1},\ldots,\theta_{R-1}$ are modelled as free parameters.

The ELBO has the same form as in~\autoref{ELBO_LVAE} except that
\begin{align}
\begin{aligned}
        \mathbb{E}_{q_{\phi}(Z \mid Y)}[\log p_{\psi}(Y \mid &Z)] =\\& \sum_{n=1}^{N} \sum_{d\in O_n}  \mathbb{E}_{q_{\phi}\left(\boldsymbol{z}_{n} \mid \boldsymbol{y}_{n}^o\right)}\left[\log p_{\psi}\left(y_{n d} \mid \bm{\gamma}_{nd} \right)\right]. 
        \label{eq:reconstruction}
\end{aligned}
\end{align}
In \autoref{eq:reconstruction}, $O_n$ corresponds to the index set of the observed variables of $\boldsymbol{y}_{n}$, and $\boldsymbol{y}_{n}^o$ represents the observed part of $\boldsymbol{y}_{n}$, i.e., a slice of $\boldsymbol{y}_{n}$ that only contains the elements index by $O_n$. The KL divergence is computed as in the case of exact inference or as in~\autoref{eq:minibatchbound} in mini-batch compatible SGD.

\subsection{Predictive distribution} 
The trained model with parameters $\phi, \psi, \theta$ can be used for making future predictions given the beginning of a sequence as well as for imputing the missing values for a given data sample. Using the learned variational approximation $q_{\phi}(Z \mid Y, X)$ in place of the intractable true posterior $p_{\omega}(Z \mid Y, X)$, the predictive distribution can be approximated as 
\begin{align*}
\begin{aligned}
&p_{\omega}\left(\boldsymbol{y}_{*} \mid \boldsymbol{x}_{*}, Y, X\right) \approx \\ 
&\int_{\boldsymbol{z}_{*}, Z} p_{\psi}\left(\boldsymbol{y}_{*} \mid \boldsymbol{z}_{*}\right) p_{\theta}\left(\boldsymbol{z}_{*} \mid \boldsymbol{x}_{*}, Z, X\right) 
q_{\phi}(Z \mid Y, X) d \boldsymbol{z}_{*} d Z \\
&=\int_{\boldsymbol{z}_{*}} \prod_{d=1}^{D} p_{\psi}\left(y_{* d} \mid \boldsymbol{\gamma}_{*d}) 
        \right) \prod_{l=1}^{L} \mathcal{N}\left(z_{* l} \mid \mu_{* l}, \sigma_{* l}^{2}\right) d \boldsymbol{z}_{*},
\end{aligned}
\end{align*}

where $\boldsymbol{y}_{*}$, $\boldsymbol{x}_{*}$, and $\boldsymbol{z}_{*}$ are the unseen test data, its associated covariates, and its corresponding latent embeddings respectively. As stated in \cite{ramchandran2021longitudinal}, the means of the latent embeddings are $\mu_{* l}=K_{\boldsymbol{x}_{*} X}^{(l)} \Sigma_{l}^{-1} \overline{\boldsymbol{\mu}}_{l}$ and its variances are $\sigma_{* l}^{2}=k_{l}\left(\boldsymbol{x}_{*}, \boldsymbol{x}_{*}\right)-K_{\boldsymbol{x}_{*} X}^{(l)} \Sigma_{l}^{-1} K_{X \boldsymbol{x}_{*}}^{(l)}+ K_{\boldsymbol{x}_{*} X}^{(l)} \Sigma_{l}^{-1} W_{l} \Sigma_{l}^{-1} K_{X \boldsymbol{x}_{*}}^{(l)}+\sigma_{z l}^{2}$, where $\bar{\boldsymbol{\mu}}_{l}$ and $W_l$ contain the encoded means and variances for all the $N$ training data points, and $\Sigma_l = \sum_{r=1}^R K_{XX}^{(l,r)} + \sigma^2_{zl} I_N$.

\section{Experiments}
\label{sec:experiments}

The ability of our model to effectively learn the underlying heterogeneous data distribution is evaluated through the model's performance in imputing missing values in the training data as well as in the unseen test data, and through the model's ability in predicting future observations. We demonstrate the efficacy of our model on two modified MNIST datasets to simulate temporal and longitudinal datasets as well as on a real clinical dataset from the Parkinson’s Progression Markers Initiative (PPMI). We compare our model's performance against HI-VAE \cite{nazabal2020handling} and L-VAE \cite{ramchandran2021longitudinal}. HI-VAE is one of the state-of-the-art VAE-based models in heterogeneous data imputation. However, it does not have explicit support for longitudinal data. On the other hand, L-VAE has demonstrated its effectiveness to successfully model longitudinal data. However, L-VAE makes the strong assumption that the data is Gaussian distributed. We demonstrate how an appropriate choice of likelihood distributions results in improved performance for temporal and longitudinal VAEs.

The additive components with different covariate functions (CF) for the additive multi-output GP prior are denoted as: ${f}_{\mathrm{se}}(\cdot)$ for squared exponential CF, ${f}_{\mathrm{ca}}(\cdot)$ for categorical CF, and ${f}_{\mathrm{ca} \times \mathrm{se}}(\cdot \times \cdot)$ for their interaction.


%
%

\begin{figure}[!t]
\includegraphics[scale=0.2]{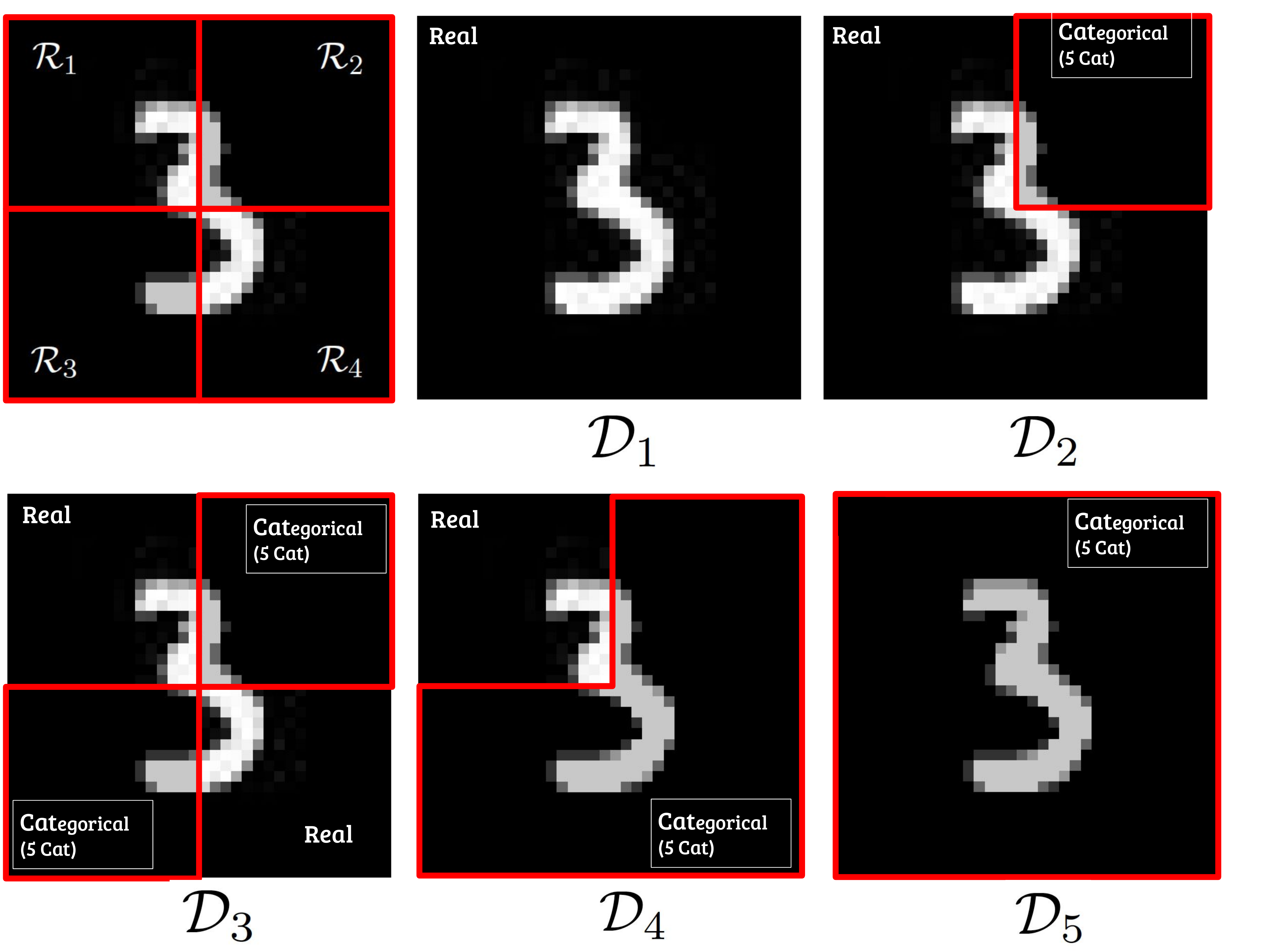} 
\centering
\caption{Heterogeneous data configuration for temporal and longitudinal MNIST dataset. Each digit comprises of 4 regions $\mathcal{R}_1, \dots, \mathcal{R}_4$. Moreover, $\mathcal{R}_i \sim \mathcal{N}$ denotes Gaussian likelihood for pixels in region $\mathcal{R}_i$, and $\mathcal{C}_j \sim \mathbb{C}_5$ denotes categorical likelihood with 5 levels for pixels in region $\mathcal{R}_j$. The datasets are simulated as: $\mathcal{D}_1 =\{\mathcal{R}_1,\mathcal{R}_2,\mathcal{R}_3,\mathcal{R}_4\}$, $\mathcal{D}_2 =\{\mathcal{R}_1,\mathcal{C}_2,\mathcal{R}_3,\mathcal{R}_4\}$, $\mathcal{D}_3 =\{\mathcal{R}_1,\mathcal{C}_2,\mathcal{C}_3,\mathcal{R}_4\}$, $\mathcal{D}_4 =\{\mathcal{R}_1,\mathcal{C}_2,\mathcal{C}_3,\mathcal{C}_4\}$ and, $\mathcal{D}_5 =\{\mathcal{C}_1,\mathcal{C}_2,\mathcal{C}_3,\mathcal{C}_4\}$}
\label{fig:4Qsetup}
\end{figure}

\subsection{MNIST Datasets}

We modified the MNIST dataset \cite{lecun1998gradient} to simulate temporal and longitudinal effects. 
To simulate heterogeneity and the effect of different observation likelihoods, we divided each image into 4 regions, $\mathcal{R}_1, \dots, \mathcal{R}_4$ and created five different datasets, $\mathcal{D}_1, \dots, \mathcal{D}_5$, as shown in \autoref{fig:4Qsetup}. 
Additionally, we masked 25\% of the pixels during the training and performed 5 independent runs with each dataset. 
As the inference network, we used two convolution layers with 144 filters, $3 \times 3$ kernel and 2 strides, following max pooling with $2 \times 2$ kernel and 2 strides and a feedforward layer of width 500. 
The latent dimensionality is 32. 
The generative network consists of two transposed convolution layers, first with 256 filters and second with $256 \times 5$ and $4 \times 4$ kernel with 2 strides followed by a fully connected layer of width 500.   
The dimension per variable in the homogenous layer, $s_{d}$, is five for both HI-VAE and HL-VAE. We use RELU activations for all layers.

\begin{figure*}[t]
\centering
\begin{subfigure}{.25\linewidth}
  \centering
  \includegraphics[width=.95\textwidth]{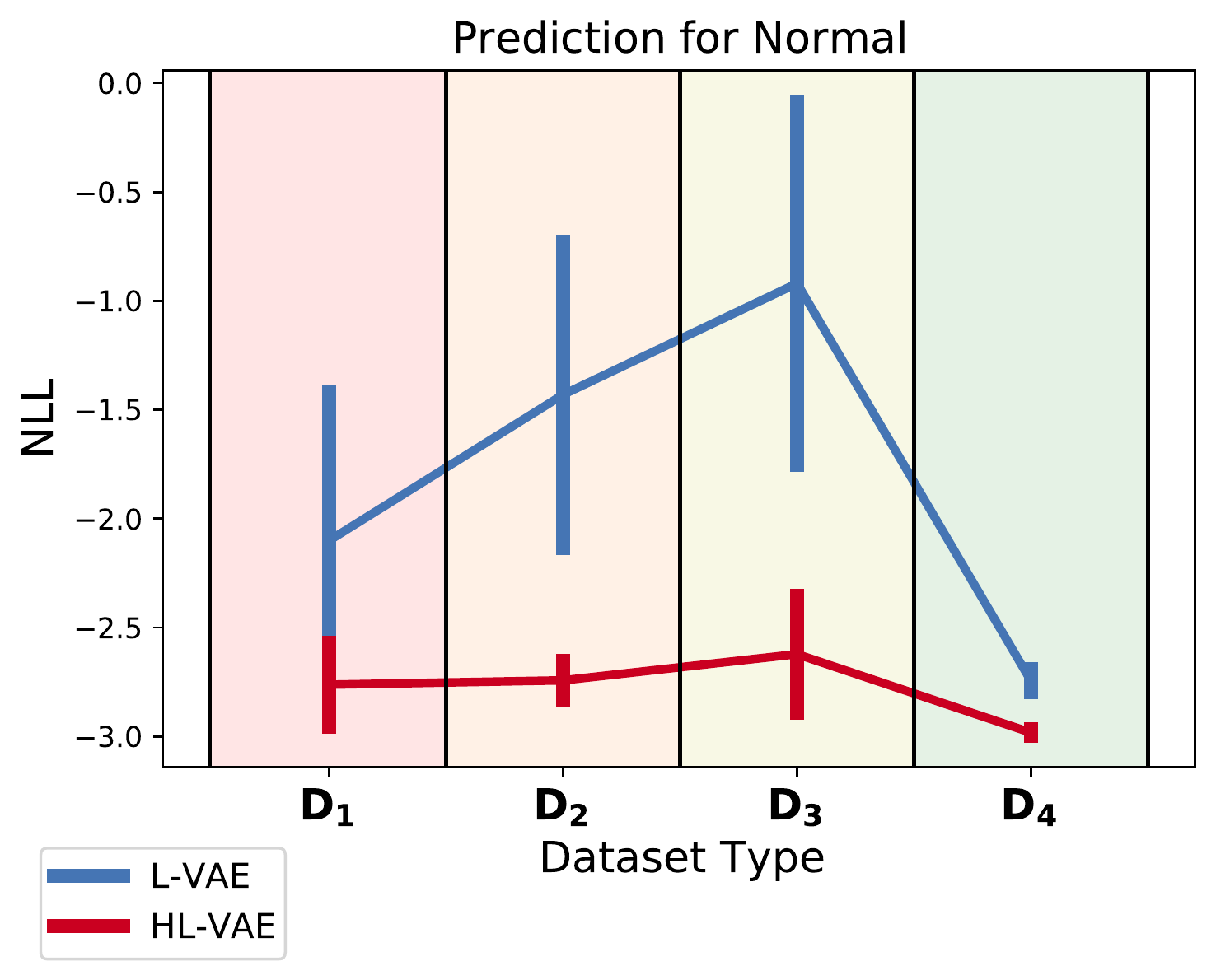}
  \caption{}
  \label{figsub:temp_mnist_a}
\end{subfigure}%
\begin{subfigure}{.5\linewidth}
  \centering
  \includegraphics[width=.95\textwidth]{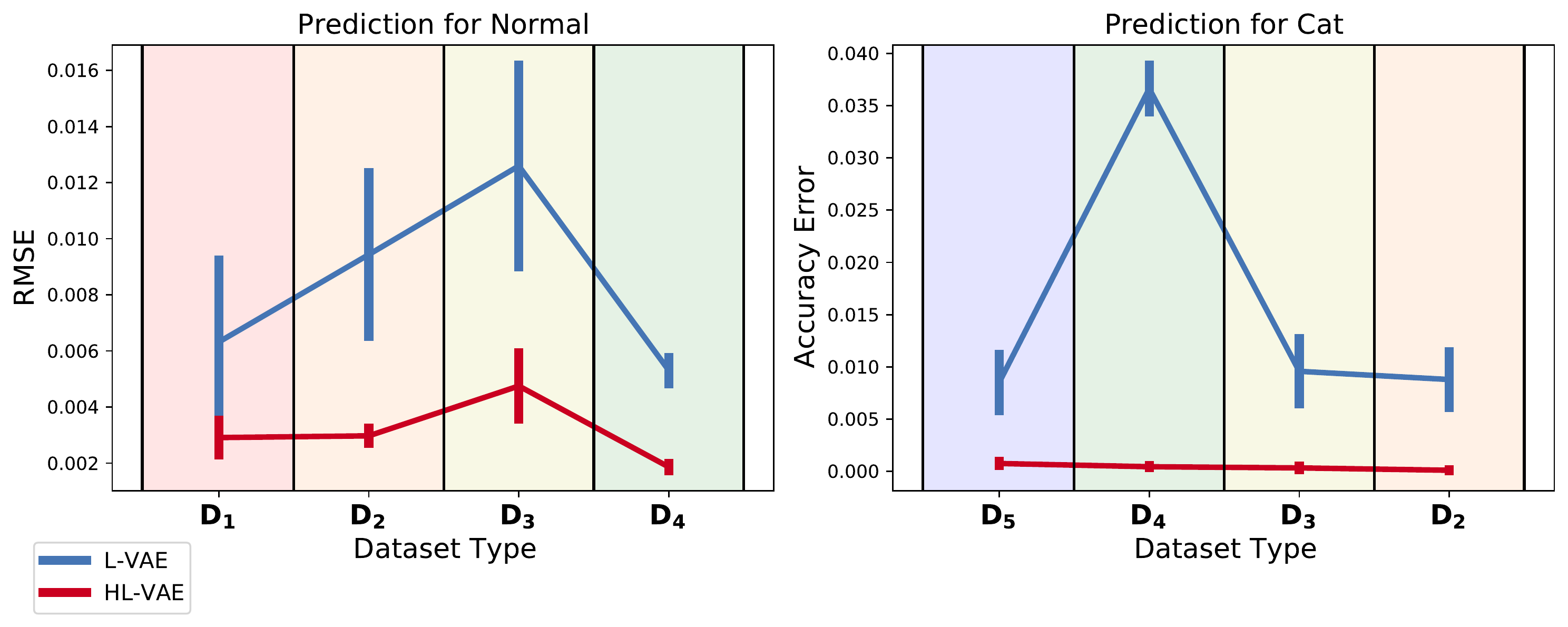}
  \caption{}
   \label{figsub:temp_mnist_b}
\end{subfigure}%
\centering
\caption{Comparison of unseen test digit prediction from auxiliary covariates of HL-VAE and L-VAE (lower is better) for temporal MNIST datasets in terms of (a) NLL for real-valued sections of the digits of $\mathcal{D}_1,\ldots,\mathcal{D}_4$ and (b) RMSE for real values (for $\mathcal{D}_1,\ldots,\mathcal{D}_4$) and accuracy error for categorical values (for $\mathcal{D}_5,\ldots,\mathcal{D}_2$). The datasets are ordered in descending order of number of variables for corresponding variable type. Vertical bars correspond to standard deviation across 5 runs.
}
\label{fig:TempMNISTResults}
\end{figure*}

\paragraph{Temporal setting} A digit is manipulated by a rotation around the centre, a diagonal shift, and a change in the intensity correlated through a time covariate. 
The manipulated MNIST images correspond to $Y$ in the dataset and the rotation, shift, intensity, and time correspond to the $X$ values (covariates)~\cite{ramchandran2022learning}. The training, validation and test datasets consist of 4000, 400 and 400 observations of a digit '3'. 
We compared L-VAE and HL-VAE in unseen digit prediction based on covariates.
The experiment results are shown in\ \autoref{fig:TempMNISTResults}. HL-VAE outperforms L-VAE in terms of negative log-likelihood (NLL) and prediction error. The accuracy for L-VAE was obtained by discretisation of the predicted continuous values. 

\paragraph{Longitudinal setting} We modified the simulated Health MNIST data used in \cite{ramchandran2021longitudinal}. 
The dataset aims to mimic medical data, where subjects (i.e., MNIST digits) with disease rotate in a sequence of 20 rotations depending on the time to disease diagnosis. Age-related effects are applied by shifting the digits towards the right corner over time. 
We evaluated the models in terms of two prediction tasks for longitudinal MNIST datasets. 
First, we compared L-VAE and HL-VAE in terms of longitudinal data prediction and evaluated the performance of predicting completely unobserved future time points given some new covariate information. 
As a second step, we compared two variations of HI-VAE, namely Gaussian prior HI-VAE and Gaussian mixture (GM) prior HI-VAE, and HL-VAE. Moreover, we also compared L-VAE and HL-VAE in terms of missing value prediction in the test split. We predicted the missing values from the observed measurements.
\begin{figure*}[!t]
\centering
\begin{subfigure}{.45\linewidth}
  \centering
  \includegraphics[width=.95\textwidth]{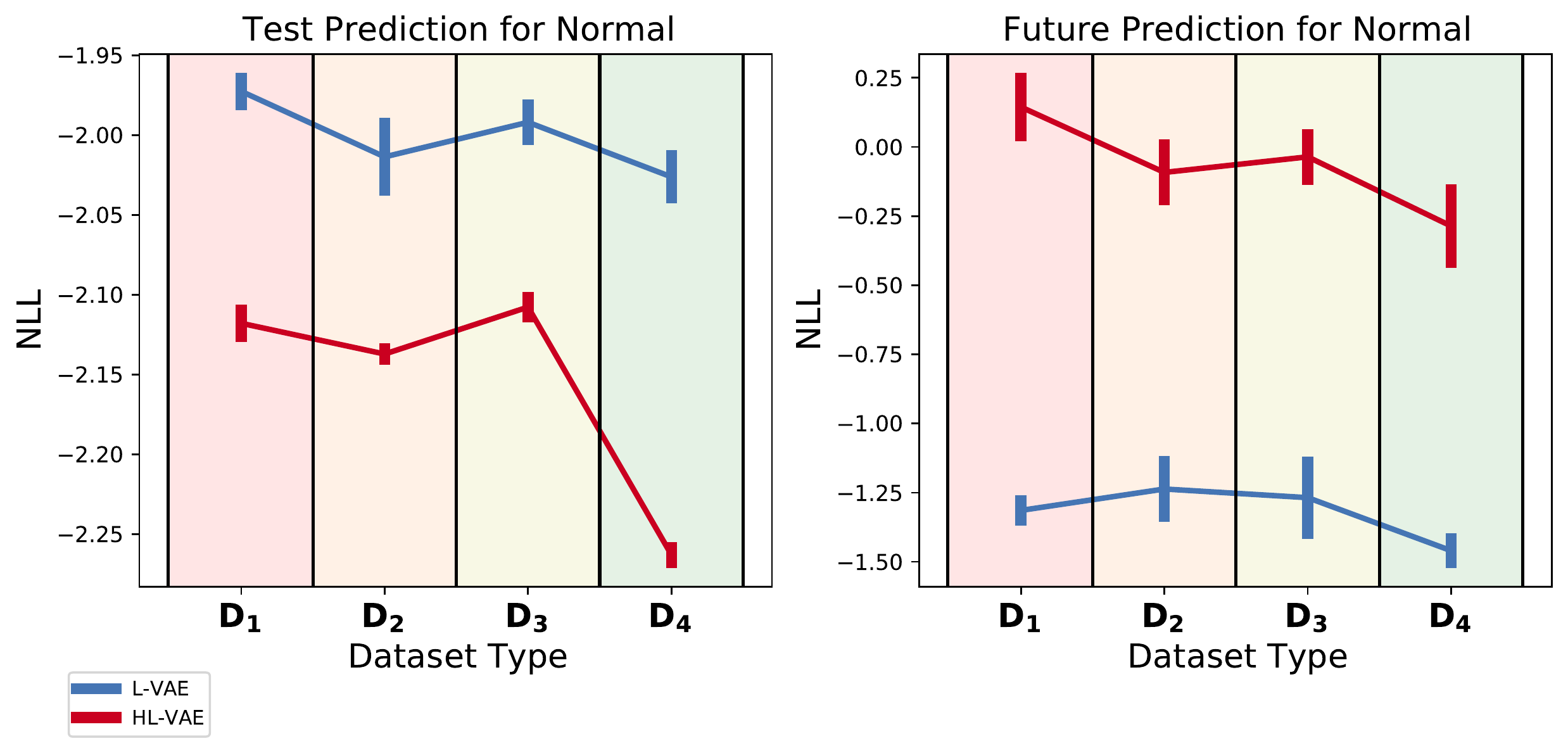}
  \caption{}
  \label{figsub:mnist_a}
\end{subfigure}%
\begin{subfigure}{.45\linewidth}
  \centering
  \includegraphics[width=.95\textwidth]{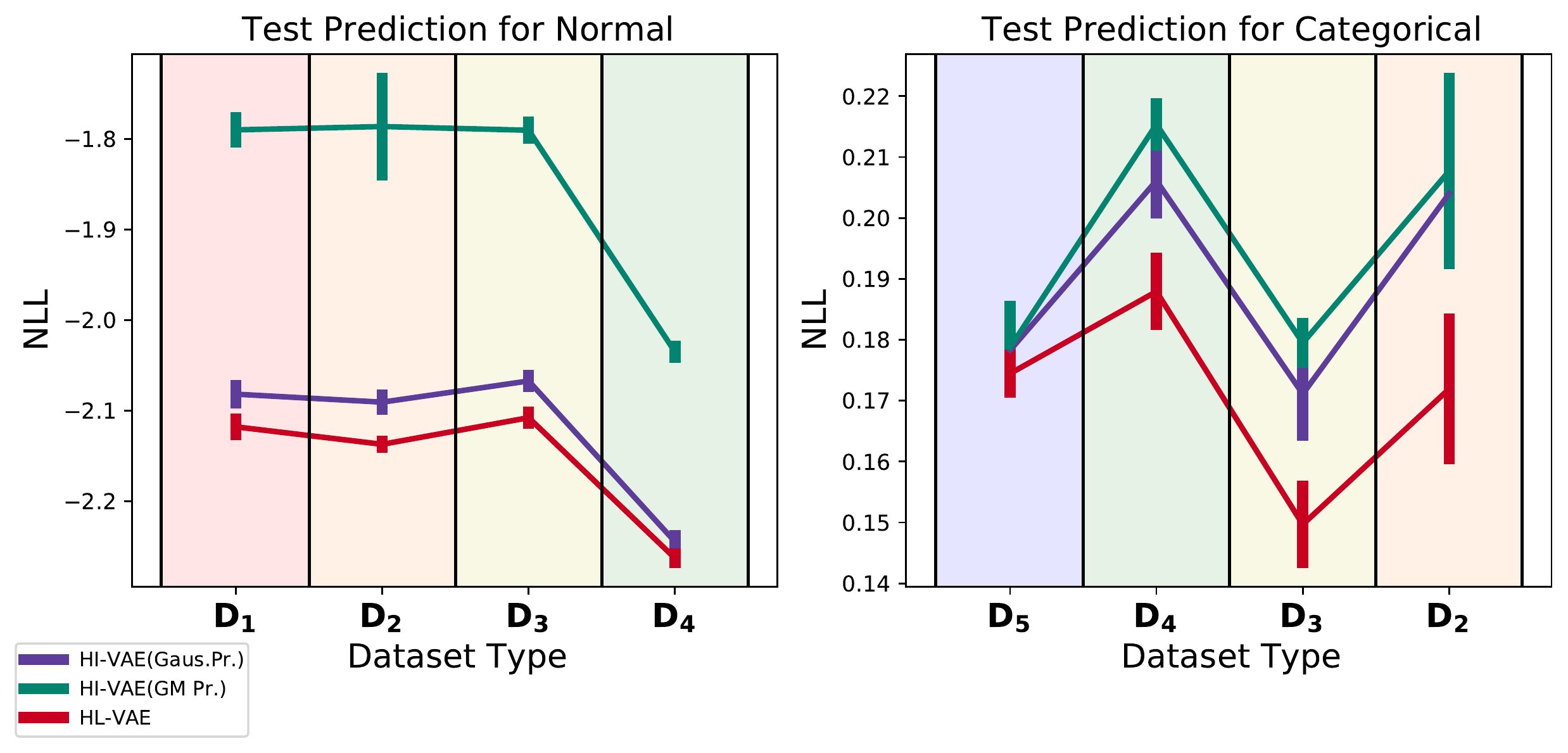}
  \caption{}
   \label{figsub:mnist_b}
\end{subfigure}%
\centering
\caption{Comparison of HL-VAE with other models for longitudinal MNIST datasets in terms of NLL (lower is better). a) Missing value prediction in test datasets and prediction of unseen future visits for L-VAE and HL-VAE with datasets $\mathcal{D}_1,\ldots,\mathcal{D}_4$ for the real-valued data (regions). b) Missing value prediction in test datasets for HI-VAE (with Gaussian and GM prior) and HL-VAE for categorical ($\mathcal{D}_5,\ldots,\mathcal{D}_2$) (left) and real-valued data ($\mathcal{D}_1,\ldots,\mathcal{D}_4$) (right).}
\label{fig:MNISTHIVAE}
\end{figure*}

\begin{table}[]
\begin{center}
\caption{Categorical (accuracy error by discretisation of continuous predictions) and real-valued variable (RMSE) errors for future prediction and predicting missing values in longitudinal MNIST test splits for L-VAE and HL-VAE.}
\label{tbl:mnist}
\centering
\resizebox{\linewidth}{!}{%
\begin{tabular}{lcccc}
    \toprule
    &\multicolumn{2}{c}{{\bfseries Categorical} (Accuracy Error $\downarrow$)} & \multicolumn{2}{c}{{\bfseries Normal} (RMSE $\downarrow$)}\\
    \cmidrule(lr){2-3}\cmidrule(lr){4-5}
    \multicolumn{1}{c}{\textbf{Dataset}} & L-VAE & HL-VAE& L-VAE & HL-VAE  \\ \midrule
\multicolumn{5}{c}{\textsc{Future Prediction}} \\ \cmidrule(lr){2-5}
$D_1$ (Real)&  -  &    - & $\mathbf{ 0.045 \sd{\pm     0.002}}$  &    $ 0.049 \sd{\pm     0.001} $\\ 
$D_2$ (1Q Cat)  &    $0.109 \sd{\pm    0.005} $ &    $\mathbf{ 0.084 \sd{\pm     0.001}}$&   $\mathbf{ 0.046\sd{\pm     0.001}}$  &    $ 0.049 \sd{\pm    0.001}$  \\ 
$D_3$ (2Q Cat)   &   $ 0.097 \sd{\pm     0.005}$  &   $\mathbf{  0.076 \sd{\pm    0.002}}$&  $\mathbf{  0.050 \sd{\pm      0.005}}$  &    $  0.055 \sd{\pm     0.002}$ \\ 
$D_4$ (3Q Cat)  &    $0.121 \sd{\pm     0.008}$  &    $\mathbf{ 0.097 \sd{\pm     0.002}}$      &  $\mathbf{  0.032 \sd{\pm      0.004}}$  &    $  0.038 \sd{\pm     0.001 }$ \\ 
$D_5$ (All Cat) &    $0.109 \sd{\pm    0.008}$  &     $\mathbf{  0.088 \sd{\pm   0.001}}$      &- & -  \\ 
\multicolumn{5}{c}{\textsc{Missing Value Prediction}} \\ \cmidrule(lr){2-5}
$D_1$ (Real)&  -  &    -       &  $  0.033  \sd{\pm     0.002}$ & $ \mathbf{    0.021 \sd{\pm     0.001 }} $  \\ 
$D_2$(1Q Cat) &     $0.092  \sd{\pm     0.008}$ & $ \mathbf{    0.059\sd{ \pm     0.004 }} $       &     $0.032  \sd{\pm     0.002}$ & $ \mathbf{    0.021 \sd{\pm     0.001} } $  \\  
$D_3$(2Q Cat) &     $0.084  \sd{\pm     0.005}$ & $ \mathbf{    0.052 \sd{\pm     0.004 }} $ &     $0.035  \sd{\pm     0.003} $& $ \mathbf{    0.022 \sd{\pm     0.001 }} $  \\
$D_4$(3Q Cat) &     $0.101  \sd{\pm     0.011}$ & $ \mathbf{    0.066 \sd{\pm     0.003 }} $        &     $0.023  \sd{\pm     0.003}$ & $ \mathbf{    0.014 \sd{\pm     0.001 }} $  \\ 
$D_5$(All Cat) &     $0.093  \sd{\pm     0.009}$ & $ \mathbf{    0.061 \sd{\pm     0.002 }} $ & $-$ & $-$ \\ \bottomrule 
\end{tabular}}
\end{center}
\end{table}

In order to compare L-VAE with HL-VAE, we used the same additive components in both models, namely $f_{\mathrm{ca}}(\mathrm{id})+f_{\mathrm{se}}(\mathrm{age})+f_{\mathrm{ca}\times \mathrm{se}}(\mathrm{id} \times \mathrm{age})+f_{\mathrm{ca} \times \mathrm{se}}(\mathrm{sex} \times \mathrm{age})+f_{\mathrm{ca} \times \mathrm{se}} (\mathrm{diseasePresence} \times \mathrm{diseaseAge}) $. Also, we make use of the same definition for $\mathrm{diseasePresence}$ and $\mathrm{diseaseAge}$ as \cite{ramchandran2021longitudinal}.
The comparison of these two methods in terms of error is given in\ \autoref{tbl:mnist} and comparison in terms of predictive NLL is given in\ \autoref{figsub:mnist_a}. 
HL-VAE outperforms L-VAE in all cases except the predictive NLL and error for the normally distributed variables of the future predictions. 

Because HI-VAE cannot process longitudinal data, we only compared it for missing value estimation.
As shown in\ \autoref{figsub:mnist_b}, HL-VAE outperforms both variations of HI-VAE in all cases. HL-VAE also has a better performance in terms of RMSE.

\begin{figure*}[!ht]
\centering
    \includegraphics[width=\textwidth]{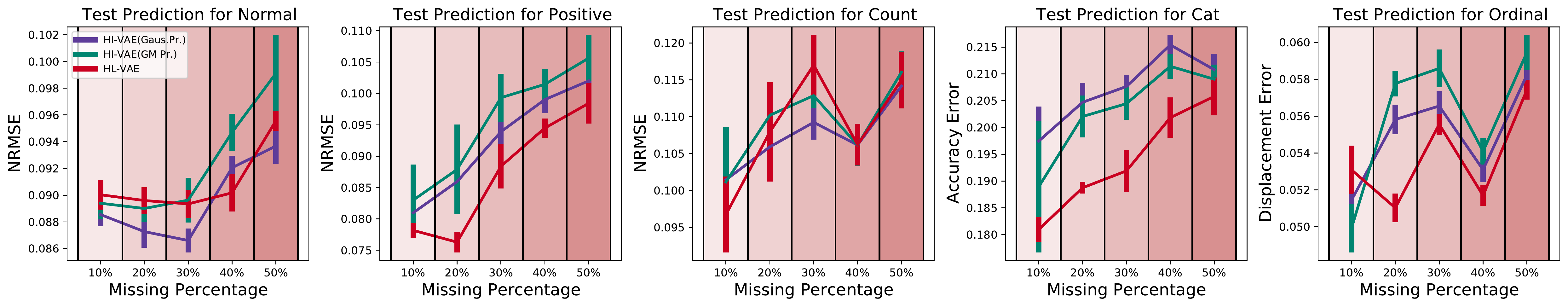}
\caption{Comparison of  the missing value prediction errors in test splits for HI-VAE with Gaussian prior, HI-VAE with Gaussian mixture prior, and HL-VAE on PPMI dataset with various missing ratios, ranging from 10\% to 50\% for all of the five likelihoods.}
\label{fig:PPMI_results_HIVAE1}
\end{figure*}

\begin{table}[]
\begin{center}
\caption{Comparison in terms of predictive NLL and NRMSE for missing time point prediction and predicting missing values in PPMI test splits for L-VAE and HL-VAE. }
\label{tbl:ppmi}
\centering
\resizebox{\linewidth}{!}{%
\begin{tabular}{lcccc}
    \toprule
    &\multicolumn{2}{c}{{\bfseries Normal} (NLL $\downarrow$)} & \multicolumn{2}{c}{{\bfseries Not Categorical} (NRMSE $\downarrow$)}\\
    \cmidrule(lr){2-3}\cmidrule(lr){4-5}
    \multicolumn{1}{c}{\textbf{Dataset}} & L-VAE & HL-VAE& L-VAE & HL-VAE  \\ \midrule
\multicolumn{5}{c}{\textsc{Missing Time Point Prediction}} \\ \cmidrule(lr){2-5}
$10\%$ Missing &  $\mathbf{     3.01  \sd{\pm      0.36}}$ &    $3.25 \sd{\pm       0.11}$  & $0.099 \sd{\pm      0.003}$ & $ \mathbf{    0.086 \sd{\pm     0.002 }} $   \\ 
$20\%$ Missing &  $\mathbf{     2.90  \sd{\pm      0.24 }}$&   $3.13 \sd{\pm     0.13}$  & $0.097  \sd{\pm     0.001}$ & $ \mathbf{    0.086 \sd{\pm     0.003} } $   \\ 
$30\%$ Missing &    $ 3.42  \sd{\pm      0.69}$ & $ \mathbf{    2.99 \sd{\pm      0.03 }} $  & $0.100  \sd{\pm     0.001}$ & $ \mathbf{    0.091 \sd{\pm     0.002 }} $  \\ 
$40\%$ Missing &    $ 4.80  \sd{\pm      0.98}$ & $ \mathbf{    3.15 \sd{\pm      0.05 }} $ & $0.102  \sd{\pm     0.001}$ & $ \mathbf{    0.088 \sd{\pm     0.004 }} $  \\ 
$50\%$ Missing &    $ 6.26  \sd{\pm      0.57}$ & $ \mathbf{    3.18 \sd{\pm      0.09 }} $ & $0.105  \sd{\pm     0.002}$ & $ \mathbf{    0.094 \sd{\pm     0.002 }} $  \\ 
\multicolumn{5}{c}{\textsc{Missing Value Prediction}} \\ \cmidrule(lr){2-5}
$10\%$ Missing &     $3.44  \sd{\pm       0.79}$ & $ \mathbf{    3.21 \sd{\pm      0.14 } }$ & $0.093  \sd{\pm      0.004}$ & $ \mathbf{    0.079 \sd{\pm     0.003 }} $  \\ 
$20\%$ Missing &     $3.22  \sd{\pm      0.49}$ & $ \mathbf{    3.06 \sd{\pm      0.04 }} $ & $0.091  \sd{\pm     0.002}$ & $ \mathbf{    0.080 \sd{\pm     0.003 }} $ \\ 
$30\%$ Missing &     $3.59  \sd{\pm      0.89}$ & $ \mathbf{    3.40 \sd{\pm      0.03 }} $ & $0.094  \sd{\pm     0.002}$ & $ \mathbf{    0.087 \sd{\pm     0.003 }} $  \\ 
$40\%$ Missing &     $4.91  \sd{\pm      1.11}$ & $ \mathbf{    3.10 \sd{\pm      0.05 }} $ & $0.095  \sd{\pm     0.001}$ & $ \mathbf{    0.085 \sd{\pm     0.001 }} $ \\ 
$50\%$ Missing &     $6.24  \sd{\pm      0.76}$ & $ \mathbf{    3.29 \sd{\pm      0.10 }} $ & $0.099  \sd{\pm     0.001}$ & $ \mathbf{    0.091 \sd{\pm     0.003 }} $  \\ \bottomrule 
\end{tabular}}
\end{center}
\end{table}

\subsection{Data from the Parkinson's Progression Markers Initiative}

This dataset (PPMI) is obtained from a large-scale, open access, international database containing multiple Parkinson's disease (PD) cohorts with longitudinal multimodal observational studies \cite{marek2011parkinson}. We used a curated dataset from the database that consists of demographic information, motor/non-motor assessments, cognitive tests, DaTSCAN, cerebrospinal fluid results, and biospecimen as well as genetics information. 
The dataset consists of measurements obtained from 545 participants who were monitored over a period of five years on 80 heterogeneous metrics, i.e., \ samples taken annually from the first visit/baseline to year five of the study, where 371 of the participants are in the PD group and 174 of them are the healthy control group.  
In addition to measurements, we used the information about the participants and their visits, such as id, gender, age at visit, the number of months that passed from the diagnosis, the duration of symptomatic therapy for those who receive, and participant group, as covariate information.
The dataset has a total of 3114 measurements with 5.5\% of actual missing values. 
We randomly selected 387, 91 and 67 participants and their associated 2212, 518 and 384 measurements as the training, test and validation sets, respectively. Moreover, we included two random visits of each participant from the test set in the training dataset to facilitate the prediction of future observations.

Hence, our dataset comprised of 80-dimensional feature vectors consisting of 8 Gaussian, 12 log-normal, 12 Poisson, 12 ordinal logit with various levels, and 36 categorical logit model with various numbers of categories.
We generated five different datasets by removing completely at random 10\% to 50\% of the data. 
We also evaluated the models 
in terms of two prediction tasks with five independent runs each. We made two random visit measurements of each participant in the test split known to the model as a part of the training split to predict the remaining (unseen) visits. 
The performance is evaluated on the artificial missing values that we introduced. 
We used the normalised root mean squared error (NRMSE) for numerical values, accuracy error for categorical values, and displacement error for ordinal values, as in \cite{nazabal2020handling}. 
For PPMI, we used a multi-layer perceptron network with a single hidden layer and ReLU activations for both inference and generative networks. The width of the hidden layer and the number of dimensions on the latent space are 50 and 8, respectively. The dimension per variable in the homogenous layer, $s_{d}$, for HI-VAE and HL-VAE is five. 

Since L-VAE cannot handle categorical information, we included only 44 non-categorical measurements. 
On the other hand, we used all the measurements for the training of HL-VAE, thereby making use of all available information. However, the comparison of the two models is performed only on the 44 non-categorical measurements. The predictive NLL and error comparisons are listed in\ \autoref{tbl:ppmi}. HL-VAE performs better in almost all predictive tasks. The ability to include the categorical measurements increased the performance of the numerical measurement predictions. 
Additionally, as the missing percentage increases, the predicted log-likelihoods decreases for L-VAE while for HL-VAE the decrease is considerably smaller. 
The comparison of HI-VAE and HL-VAE in terms of predictive error is shown in\ \autoref{fig:PPMI_results_HIVAE1}. 
In terms of predicting the missing values in the test split, HL-VAE has lower errors for all data types except for count values. 


\section{Discussion}
\label{sec:discussion}
Longitudinal datasets, such as clinical trial data, contain vital information that is important to appropriately model while taking into account the intricacies such data introduces. 
In this paper, we proposed an extension to the VAE that can analyse high-dimensional, heterogeneous temporal and longitudinal data. 
Moreover, we empirically showed that our model is adept at imputing missing values as well as performing future predictions that outperform competing approaches. 
We demonstrated the benefits of choosing appropriate likelihood models for the data while taking the temporal aspect of the data into account. 
Since our model makes more appropriate assumptions, we believe our work will be an important contribution to the analysis of temporal and longitudinal datasets.



\bibliographystyle{IEEEtran}
\bibliography{conference_101719}

\end{document}